\documentclass{article} 
\usepackage{iclr2017_conference,times}
\usepackage{hyperref}
\usepackage{url}

\usepackage{algorithm}
\usepackage{algorithmic}
\usepackage{graphicx}
\usepackage{color}
\usepackage{amsmath,amssymb}

\title{Transformation-Based Models of Video\\Sequences}

\author{Joost van Amersfoort \thanks{Work done as part of internship with FAIR}, Anitha Kannan, Marc'Aurelio Ranzato, \\
  \textbf{Arthur Szlam, Du Tran \& Soumith Chintala} \\
  Facebook AI Research \\
  \texttt{joost.van.amersfoort@cs.ox.ac.uk}, \\
  \texttt{\{akannan, ranzato, aszlam, trandu, soumith\}@fb.com}}

\begin{document}

\maketitle

\begin{abstract}
  In this work we propose a simple unsupervised approach for next frame prediction in
  video. Instead of directly predicting the pixels in a frame given past
  frames, we predict the transformations needed for generating the next
  frame in a sequence, given the transformations of the past frames. This leads
  to sharper results, while using a smaller prediction model.
  
  In order to enable a fair comparison between different video frame prediction
  models, we also propose a new evaluation protocol. We use generated frames as
  input to a classifier trained with ground truth sequences. This criterion
  guarantees that models scoring high are those producing sequences which
  preserve discriminative features, as opposed to merely penalizing any
  deviation, plausible or not, from the ground truth. Our proposed approach
  compares favourably against more sophisticated ones on the UCF-101 data
  set, while also being more efficient in terms of the number of parameters and
  computational cost.
\end{abstract}

\section{Introduction}
There has been an increased interest in unsupervised learning of representations
from video
sequences~\citep{mathieu2015deep,srivastava2015unsupervised,vondrick2016generating}.
A popular formulation of the task is to learn to predict a small number of
future frames given the previous K frames; the motivation being that predicting
future frames requires understanding how objects interact and what plausible
sequences of motion are. These methods directly aim to predict pixel values,
with either MSE loss or adversarial loss.

\begin{figure}[t]
  \begin{center}
    \centerline{\includegraphics[width=0.5\columnwidth]{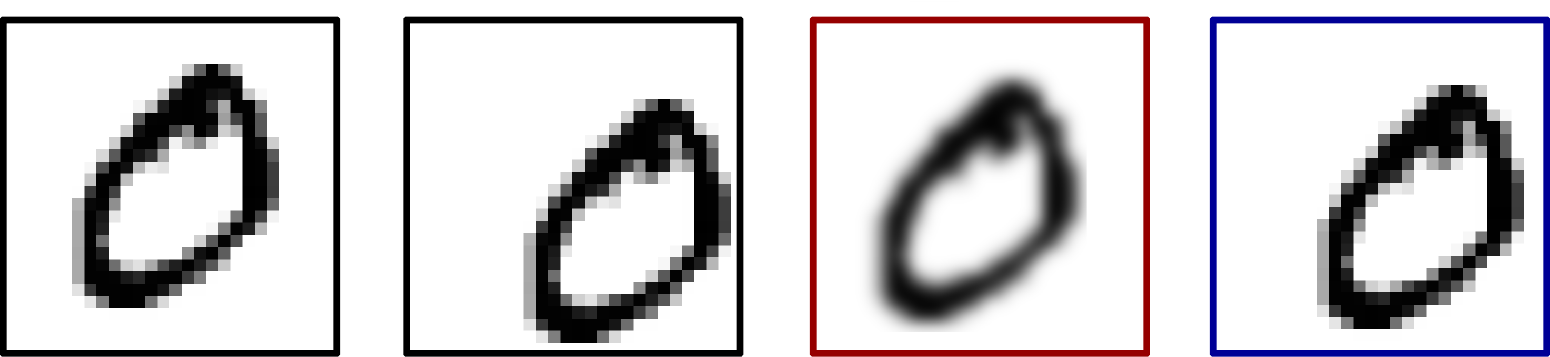}}
    \caption{Motivating toy example. From left to right: the first digit shows what
      the model is conditioned upon, the second digit shows the frame we would like
      to predict at the next time step, the third digit shows the blurry prediction
      if we were to minimize MSE in pixel space, the last digit shows the prediction
      when minimizing MSE in the space of transformations. While the two models may
      have the same MSE in pixel space, the transformation-based model generates
      much sharper outputs. Although the motion is different than the ground truth
      (second digit), it is still a plausible next frame to the conditioned frame.
      In practice, the input is a sequence of consecutive frames.}
    \label{demo}
  \end{center}
\end{figure}

In this paper, we take a different approach to the problem of next frame
prediction. In particular, our model operates in the space of transformations
between frames, directly modeling the source of variability. We exploit the
assumption that the transformations of objects from frame to frame should be
smooth, even when the pixel values are not. Instead of predicting pixel values,
we directly predict how objects transform. The key insight is that while there
are many possible outputs, predicting one such transformation will yield motion
that may not correspond to ground truth, yet will be realistic; see fig.
~\ref{demo}. We therefore propose a {\em transformation-based} model that
operates in the space of affine transforms. Given the affine transforms of a few
previous frames, the model learns to predict the local affine transforms that
can be deterministically applied on the image patches of the previous frame to
generate the next frame. The intuition is that estimation errors will lead to a
slightly different yet plausible motion. Note that this allows us to keep using
the MSE criterion, which is easy to optimize, as long as it is in transformation
space. No blur in the pixel space will be introduced since the output of the
transformation model is directly applied to the pixels, keeping sharp edges
intact. Refer to fig.~\ref{prediction_results} and our online material
\footnote{see: \url{https://y0ast.github.io/Transformation-Based-Models-of-Video-Sequences/GenerationBenchmark/}} for examples.

\looseness=-1
The other contribution of this work is the evaluation protocol. Typically,
generative models of video sequences are evaluated in terms of MSE in pixel
space~\citep{srivastava2015unsupervised}, which is not a good choice since this
metric favors blurry predictions over other more realistic looking options that
just happen to differ from the ground truth. Instead, we propose to feed the
generated frames to a video classifier trained on ground truth sequences. The
idea is that the less the classifier's performance is affected by the generates
frames the more the model has preserved distinctive features and the more the generated
sequences are plausible. Regardless of whether they resemble the actual ground
truth or not. This protocol treats the classifier as a black box to measure how
well the generated sequences can serve as surrogate for the truth sequence for
the classification task. In this paper we will validate our assumption that
motion can be modelled by local affine transforms, after which we will compare
our method with networks trained using adversarial training and simple
regression on the output frame, using both this new evaluation protocol and by
providing samples for qualitative inspection.

Our experiments show that our simple and efficient model outperforms
other baselines, including much more sophisticated models, on benchmarks on the
UCF-101 data set~\citep{soomro2012ucf101}. We also provide qualitative comparisons
to the moving MNIST digit data set~\citep{srivastava2015unsupervised}.

\subsection{Related Work}

Early work on video modeling focused on predicting small
patches~\citep{memisevic_nips14,srivastava2015unsupervised}; unfortunately,
these models have not shown to scale to the complexity of high-resolution
videos. Also these models require a significant amount of parameters and
computational power for even relatively simple data.

In \cite{ranzato2014video}, the authors circumvented this problem by quantizing
the space of image patches. While they were able to predict a few
high-resolution frames in the future, it seems dissatisfying to impose such a
drastic assumption to simplify the prediction task.

\cite{mathieu2015deep} recently proposed to replace MSE in pixel
space with a MSE on image gradients, leveraging prior domain knowledge, and
further improved using a multi-scale architecture with adversarial
training~\citep{adversarialnet}. While producing better results than earlier
methods, the models used require a very large amount of computational power. We
make an explicit comparison to this paper in the experiments
section~\ref{sec:experiments}.

\looseness=-1
\cite{jia2016dynamic} describe a model where filters are learned for all
locations in the input frame. The model is trained end-to-end and results on the
moving mnist dataset and a private car video dataset are shown. Even though the
paper also works on the problem of next frame prediction, it differs quite
substantially from this work. The most prominent difference is the fact that it
works in the pixelspace. Our model outputs solely the affine transformation,
requiring very few parameters to do this.

A recent strong result is provided in~\cite{xue2016visual}. This paper describes
a model that generates videos which exhibit substantial motion using a motion
encoder, an image encoder and a cross convolution part with a decoder. This
model also focuses on directly generating the pixels; however, as opposed to
dynamic filter networks, the model is trained to generate the difference
image for the next time step. By doing this, the model makes a strong implicit
assumption that the background is uniform, without any texture, so that the
differencing operation captures only the motion for the foreground object. In
contrast, our model does not make such assumptions, and it can be applied to natural
videos.
\begin{figure*}[t] 
  \begin{center}
    \centerline{\includegraphics[width=1\columnwidth]{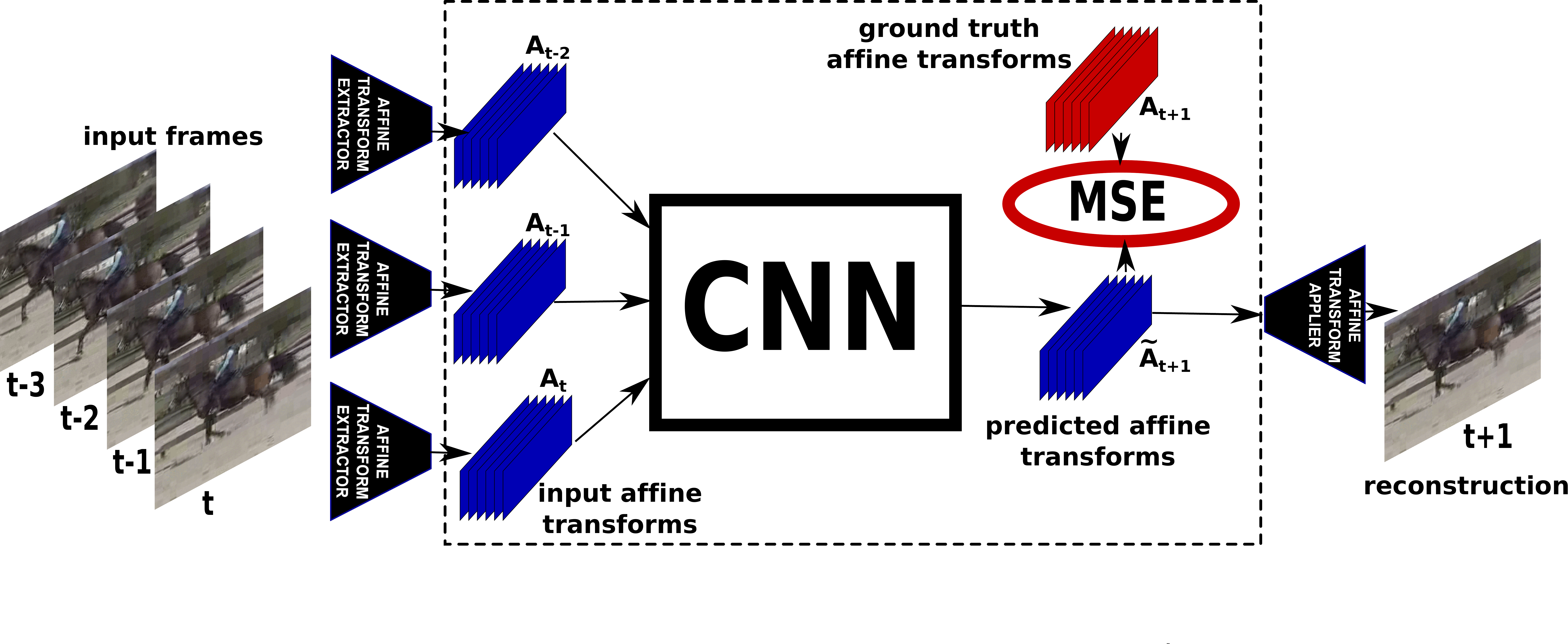}}
    \caption{Outline of the transformation-based model. The model is a CNN that
      takes as input a sequence of consecutive affine transforms between pairs of
      adjacent video frames. It predicts the affine transform between the last input
      frame and the next one in the sequence. We compute affine transforms (6
      parameters per patch) for overlapping patches of size $8\times8$ in each video
      frame. Learning operates in the space of transformations as shown inside the
      dashed box. The front-end on the left is a module that estimates the affine
      transforms between pairs of consecutive input frames. The post-processor on the
      right reconstructs a frame from the predicted set of affine transforms and it is
      only used at test time.}
    \label{model}
  \end{center}
\end{figure*}

\cite{walker2016uncertain} describe a conditional VAE model consisting of three
towers, an image tower, an encoder tower and a decoder tower. During training
the model is given an input image and a set of trajectories, it is trained to
reconstruct these input trajectories. The important difference is that during
test time, given an input image, the model simply samples from the prior
distribution over Z: the goal is to produce trajectories corresponding to that
image, that seem likely given the full data set.

In~\cite{oh2015action}, and similarly~\cite{finn2016unsupervised} for Robot
tasks and~\cite{byravan2016se3} for 3D objects, frames of a video game are predicted given an
action (transformation) taken by an agent. While the papers show great
results, the movement in a natural video cannot be described by a simple action
and these methods are therefore not widely applicable.

Perhaps most similar to our approach, \cite{patraucean2015spatio} also separate
out motion/content and directly model motion and employs the Spatial
Transformer network~\citep{jaderberg2015spatial}. The biggest difference is that
our approach is solely convolutional, which makes training fast and the
optimization problem simpler. This also allows the model to scale to larger
datasets and images, with only modest memory and computational resources. The
model directly outputs full affine transforms instead of pixels (rather than
only translations as in equation 3 in~\cite{patraucean2015spatio}).

Prior work relating to the evaluation protocol can be found
in~\cite{yan2015attribute2image}. The authors generate images using a set of
predefined attributes and later show that they can recover these using a
pretrained neural network. Our proposal extends this to videos, which is more
complicated since both appearance and motion are needed for correct
classification.

\section{Model}
\label{sec:model}

The model we propose is based on three key assumptions: 1) just estimating
object motion yields sequences that are plausible and relatively sharp, 2)
global motion can be estimated by tiling high-resolution video frames into
patches and estimating motion ``convolutionally'' at the patch level, and 3)
patches at the same spatial location over two consecutive time steps undergo a
deformation which can be well described by an affine transformation.
\begin{figure*}[ht]
  \begin{center}
    \centerline{\includegraphics[width=1\columnwidth]{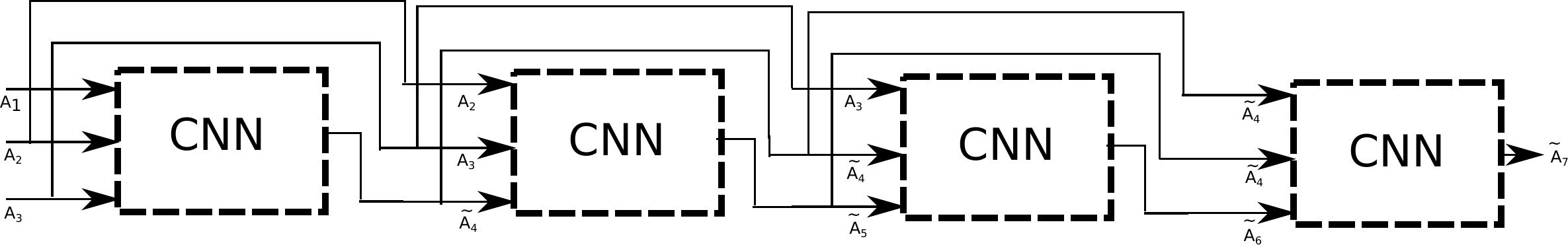}}
    \caption{Outline of the system predicting $4$ frames ahead in time. Only affine
      transforms $A_1$, $A_2$ and $A_3$ are provided, and the model predicts
      $\tilde{A}_4$, $\tilde{A}_5$, $\tilde{A}_6$ and $\tilde{A}_7$, which are used
      to reconstruct the next $4$ frames. Since affine parameters are continuous
      values and the whole chain of CNNs is differentiable, the whole unrolled
      system can be trained by back-propagation of the error. Note that CNNs all share the same parameters}
    \label{unrolling}
  \end{center}
\end{figure*}
The first assumption is at the core of the proposed method: by considering
uncertainty in the space of transformations we produce sequences that may still
look plausible. The other two assumptions state that a video sequence can be
composed by patches undergoing affine transformations. We agree that these are
simplistic assumptions, which ignore how object identity affects motion and do
not account for out of plane rotations and more general forms of deformation.
However, our qualitative and quantitative evaluation shows the efficacy of these
assumptions to real video sequence as can be seen in
section~\ref{sec:experiments} and from visualizations in the supplementary material\footnote{see: \url{https://y0ast.github.io/Transformation-Based-Models-of-Video-Sequences/ReconstructionsFromGroundTruth/} and \\ \url{https://y0ast.github.io/Transformation-Based-Models-of-Video-Sequences/GenerationBenchmark/}
}.

Our approach consists of three steps. First, we estimate affine transforms of
every video sequence to build a training set for our model. Second, we train a
model that takes the past $N$ affine transforms and predicts the next $M$ affine
transforms. Finally, at test time, the model uses the predicted affine
transforms to reconstruct pixel values of the generated sequence. We describe
the details of each phase in the following sections.

\subsection{Affine Transform Extractor}
\label{sec:ate}

Given a frame $\*x$ and the subsequent frame $\*y$, the goal of the affine
transform extractor is to learn mappings that can warp $\*x$ into $\*y$. Since
different parts of the scene may undergo different transforms, we tile $\*x$
into overlapping patches and infer a transformation for each patch. The
estimation process couples the transformations at different spatial locations
because we minimize the reconstruction error of the entire frame $\*y$, as
opposed to treating each patch independently.

Let $\*x$ and $\*y$ have size $D_r \times D_c$. Let image $\*x$ be decomposed
into a set of overlapping patches, each containing pixels from patches of size
$d_r \times d_c$ with $d_r \le D_r$ and $d_c \le D_c$. These patches are laid
out on a regular grid with stride $s_r$ and $s_c$ pixels over rows and columns,
respectively. Therefore, every pixel participates in $\frac{d_r}{s_r}
  \frac{d_c}{s_c}$ overlapping patches, not taking into account for the sake of
simplicity border effects and non-integer divisions. We denote the whole set of
overlapping patches by $\{X_k\}$, where index $k$ runs over the whole set of
patches. Similarly and using the same coordinate system, we denote by $\{Y_k\}$
the set of overlapping patches of $\*y$.

We assume that there is an affine mapping $A_k$ that maps $X_k$ to
$Y_k$, for all values of $k$. $A_k$ is a $2\times3$ matrix of free parameters
representing a generic affine transform (translation, rotation and scaling)
between the coordinates of output and input frame. Let $\tilde{Y}_{k}$ be the
transformed patches obtained when $A_k$ is applied to $X_k$. Since coordinates
overlap between patches, we reconstruct $\*y$ by averaging all predictions at
the same location, yielding the estimate $\tilde{\*y}$. The joint set of $A_k$
is then jointly determined by minimizing the mean squared reconstruction error
between $\*y$ and $\tilde{\*y}$.

Notice that our approach and aim differs from spatial transformer
networks~\citep{jaderberg2015spatial} since we perform this estimation off-line
only for the input frames, computing one transform per patch.

In our experiments, we extracted $16 \times 16$ pixel patches from the input and
we used stride $4$ over rows and columns. The input patches are then matched at
the output against smaller patches of size $8 \times 8$ pixels, to account for
objects moving in and out of the patch region.

\subsection{Affine Transform Predictor}
The affine transform predictor is used to predict the affine transforms between
the last input frame and the next frame in the sequence. A schematic
illustration of the system is shown in fig.~\ref{model}. It receives as input
the affine transforms between pairs of adjacent frames, as produced by the
affine transform extractor described in the previous section. Each transform is
arranged in a grid of size $6 \times n \times n$, where $n$ is the number of
patches in a row/column and $6$ is the number of parameters of each affine
transform. Therefore, if four frames are used to initialize the model, the
actual input consists of $18$ maps of size $n \times n$, which are the
concatenation of ${A_{t-2}, A_{t-1}, A_t}$, where $A_t$ is the collection of
patch affine transforms between frame at time $t-1$ and $t$.

The model consists of a multi-layer convolutional network without any pooling.
The network is the composition of convolutional layers with ReLU non-linearity,
computing a component-wise thresholding as in $v = \max(0, u)$. We learn the
parameters in the filters of the convolutional layers by minimizing the
mean squared error between the output of the network and the target transforms.
Notice that we do not add any regularization to the model. In particular, we
rely on the convolutional structure of the model to smooth out predictions at
nearby spatial locations.

\subsection{Multi-Step Prediction}
\label{sec:unroll}

In the previous section, we described how to predict the set of affine transforms
at the next time step. In practice, we would like to predict several time steps
in the future.
A greedy approach would: a) train as described above to minimize the prediction
error for the affine transforms at the next time step, and b) at test time,
predict one step ahead and then re-circulate the model prediction back to the
input to predict the affine transform two steps ahead, etc.
Unfortunately, errors may accumulate throughout this process because the model
was never exposed to its own predictions at training time.

The approach we propose replicates the model over time, also during training as
shown in fig.~\ref{unrolling}. If we wish to predict $M$ steps in the future, we
replicate the CNN $M$ times and pass the output of the CNN at time step $t$ as
input to the same CNN at time step $t+1$, as we do at test time. Since
predictions live in a continuous space, the whole system is differentiable and
amenable to standard back-propagation of the error. Since parameters of the CNN
are shared across time, the overall system is equivalent to a peculiar recurrent
neural network, where affine transforms play the role of recurrent states. The
experiments in section~\ref{sec:experiments} demonstrate that this method is more
accurate and robust than the greedy approach.

\subsection{Testing}
\label{sec:test}

At test time, we wish to predict $M$ frames in the future given the past $N$
frames. After extracting the $N-1$ affine transforms from the frames we
condition upon, we replicate the model $M$ times and feed its own prediction
back to the input, as explained in the previous section.

Once the affine transforms are predicted, we can reconstruct the actual pixel
values. We use the last frame of the sequence and apply the first set of affine
transforms to each patch in that frame. Each pixel in the output frame is
predicted multiple times, depending on the stride used. We average these
predictions and reconstruct the whole frame. As required, we can repeat this
process for as many frames as necessary, using the last reconstructed frame and
the next affine transform.

In order to evaluate the generation, we propose to feed the generated frames to
a trained classifier for a task of interest. For instance, we can condition the
generation using frames taken from video clips which have been labeled with the
corresponding action. The classifier has been trained on ground truth data but
it is evaluated using frames fantasized by the generative model. The performance
of the classifier on ground truth data is an upper bound on the performance of
any generative model. This evaluation protocol does not penalize any generation
that deviates from the ground truth, as standard MSE would. It instead check
that discriminative features and the overall semantics of the generated sequence
is correct, which is ultimately what we are interested in.

\section{Experiments}
\label{sec:experiments}

In this section, we validate the key assumptions made by our model and compare
against state-of-the-art generative models on two data sets. We strongly
encourage the reader to watch the short video clips in the Supplementary
Material to better understand the quality of our generations.

In section~\ref{sec:model}, we discussed the three key assumptions at the
foundations of our model: 1) errors in the transformation space look still
plausible, 2) a frame can be decomposed into patches, and 3) each patch motion
is well modeled by an affine transform. The results in the Supplementary
Material \footnote{see: \url{https://y0ast.github.io/Transformation-Based-Models-of-Video-Sequences/ReconstructionsFromGroundTruth/}
} validate assumption 2 and 3 qualitatively. Every row shows a sequence from the
UCF-101 dataset~\citep{soomro2012ucf101}. The column on the left shows the
original video frames and the one on the right the reconstructions from the
estimated affine transforms, as described in section~\ref{sec:ate}. As you can
see there is barely any noticeable difference between these video sequences,
suggesting that video sequences can be very well represented as tiled affine
transforms. For a quantitative comparison and for an assessment of how well the
first assumption holds, please refer to section~\ref{sec:ucf101}.

In the next section, we will first report some results using the toy data set of
``moving MNIST digits''~\citep{srivastava2015unsupervised}. We then discuss
generations of natural high-resolution videos using the UCF-101 dataset and
compare to current state-of-the-art methods.

\begin{figure}[t]
  \begin{center}
    \centerline{\includegraphics[width=1\columnwidth]{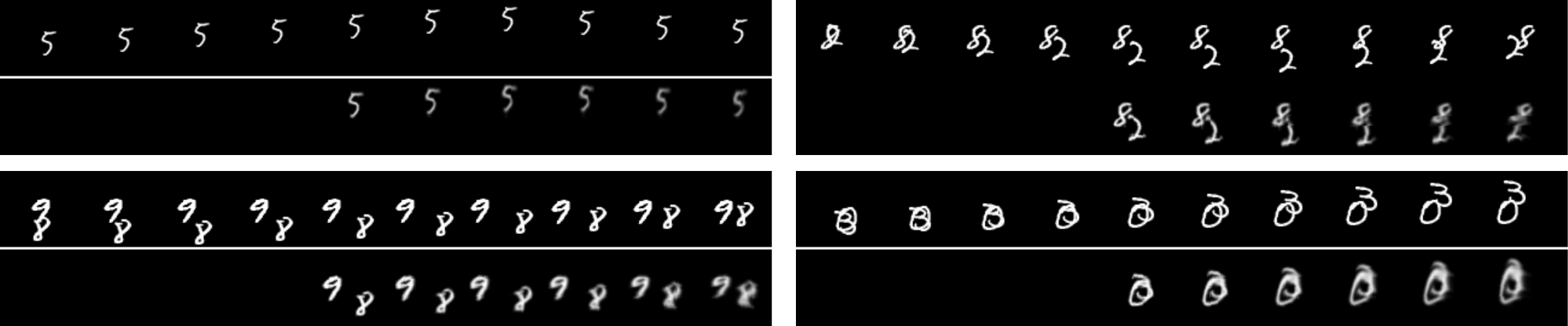}}
    \caption{Predictions of $4$ sequences from the moving MNIST dataset. The top row
      of each pair shows the ground truth frames; the first four frames are used as
      input to the model. The bottom row shows the predictions of the model.}
    \label{fig:mnist}
  \end{center}
\end{figure}

\subsection{Moving MNIST}
\label{sec:mnist}

For our first experiment, we used the dataset of moving MNIST
digits~\citep{srivastava2015unsupervised} and perform qualitative
analysis\footnote{A quantitative analysis would be difficult for this data set
  because metrics reported in the literature like
  MSE~\citep{srivastava2015unsupervised} are not appropriate for measuring
  generation quality, and it would be difficult to use the metric we propose
  because we do not have labels at the sequence level and the design of a
  classifier is not trivial.}. It consists of one or two MNIST digits, placed at
random locations and moving at constant speed inside a $64\times64$ frame. When
a digit hits a boundary, it bounces, meaning that velocity in that direction is
reversed. Digits can occlude each other and bounce off walls, making the data set
challenging.

Using scripts provided by~\cite{srivastava2015unsupervised}, we generated a
fixed dataset of 128,000 sequences and used 80\% for training, 10\% for
validation and 10\% for testing. Next, we estimated the affine transforms
between every pair of adjacent frames to a total of 4 frames, and trained a
small CNN in the space of affine transforms. The CNN has 3 convolutional
layers and the following number of feature maps: $18$, $32$, $32$, $6$. All
filters have size $3 \times 3$.

Fig.~\ref{fig:mnist} shows some representative test sequences and the model
outputs. Each subfigure corresponds to a sequence from the test set; the top row
corresponds to the ground truth sequence while the bottom row shows the
generations. The input to the CNN are three sets of affine transforms
corresponding to the first four consecutive frames. The network predicts the
next six sets of affine transforms from which we reconstruct the corresponding
frames. These results should be compared to fig. 5
in~\cite{srivastava2015unsupervised}. The
generations in fig.~\ref{fig:mnist} show that the model has potential to
represent and generate video sequences, it learns to move digits in the right
direction, to bounce them, and it handles multiple digits well except when
occluion makes inputs too ambiguous. The model's performance is analyzed
quantitatively in the next section using high resolution natural videos.

\subsection{UCF 101 data set}
\label{sec:ucf101}

The UCF-101 dataset~\citep{soomro2012ucf101} is a collection of 13320 videos of
101 action categories. Frames have size $240\times320$ pixels. We train a CNN on
patches of size $64\times64$ pixels; the CNN has $6$ convolutional layers and
the following number of feature maps: $18$, $128$, $128$, $128$, $64$, $32$,
$16$, $6$. All filters have size $3 \times 3$. The optimal number of filters has been
found using cross-validation in order to minimize the estimation error of the
affine transform parameters. Unless otherwise stated, we condition generation on
$4$ ground truth frames and we predict the following $8$ frames.

\begin{figure}[t]
  \begin{center}
    \centerline{\includegraphics[width=1\columnwidth]{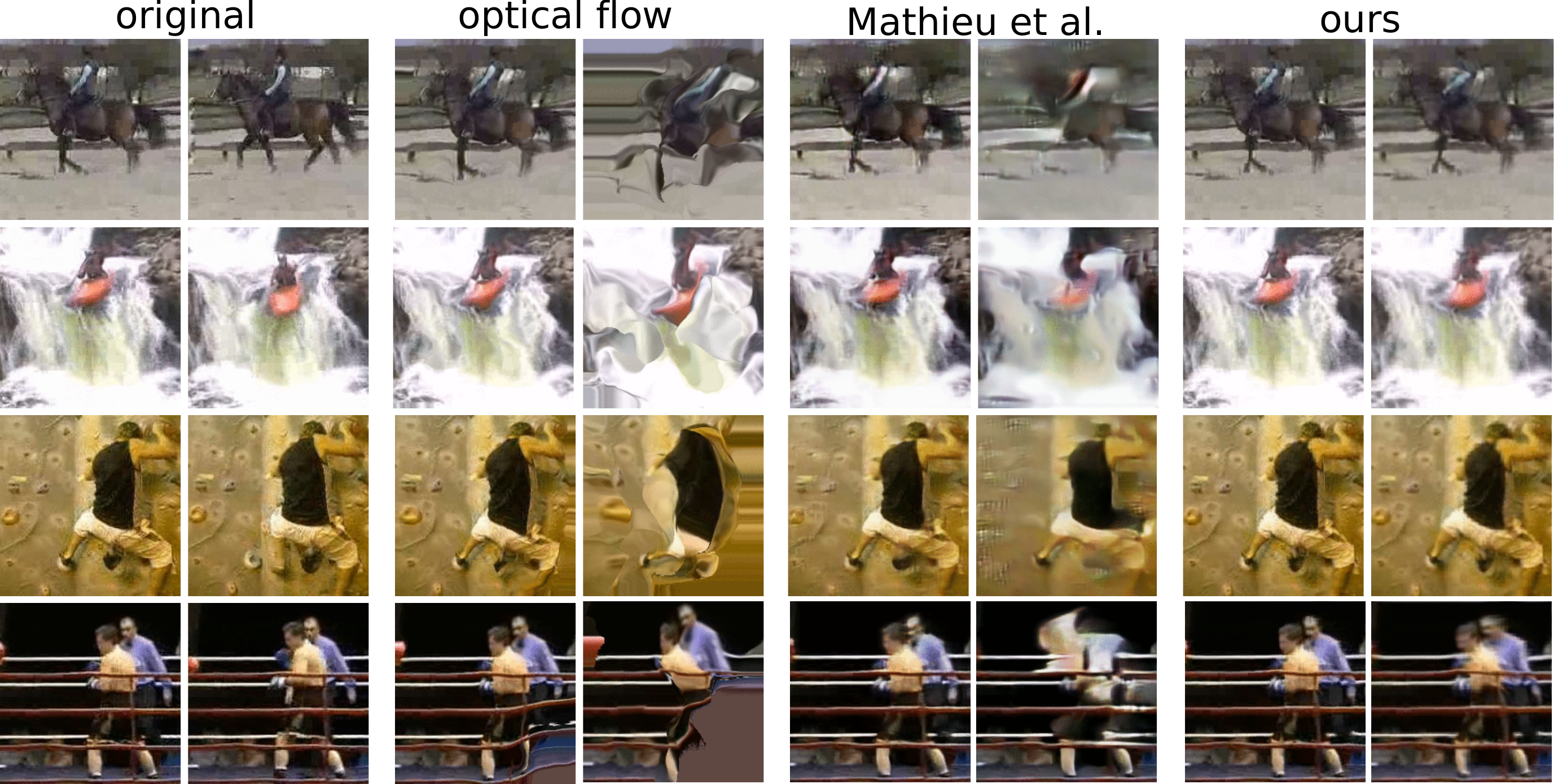}}
    \caption{Example of predictions produced by different models. Each row shows an
      example. The first two columns show the ground truth. The two frames are $4$
      time steps apart. The next two columns show predictions from a baseline model
      employing optical flow. Next, we show the prediction produced by the
      adversarially trained CNN proposed by~\cite{mathieu2015deep}. The last two
      column show the prediction produced by our affine-transformation based approach.
      All pairs in the same column group are four time steps apart. All methods were
      conditioned on the same set of 4 input frames (not shown in the figure) }
    \label{prediction_results}
  \end{center}
\end{figure}

\setlength{\tabcolsep}{4pt}
\begin{table}[bt] \centering
  \caption{Classification accuracy on UCF-101 dataset. The classifier is trained
    on the actual training video sequences, but it is tested using frames generated
    by various generative models. Each column shows the accuracy on the test set
    when taking a different number of input frames as input. Our approach maps
    $16\times16$ patches into $8\times8$ with stride $4$, and it takes $4$ frames at
    the input.}
  \label{table:ucf101}
  \begin{tabular}{l||c | c | c | c}
    Method                                                & 4 frames & 8 frames \\
    \hline \hline
    Ground truth frames                                   & 72.46    & 72.29    \\
    \hline Using ground truth affine transforms           & 71.7     & 71.28    \\
    \hline Copy last frame                                & 60.76    & 54.27    \\
    \hline Optical Flow                                   & 57.29    & 49.37    \\
    \hline \cite{mathieu2015deep}                         & 57.98    & 47.01    \\
    \hline ours - one step prediction (not unrolled)      & 64.13    & 57.63    \\
    \hline ours - four step prediction (unrolled 4 times) & 64.54    & 57.88
  \end{tabular}
\end{table} \setlength{\tabcolsep}{1.4pt}

We evaluate several models\footnote{Unfortunately, we could not compare against
  the LSTM-based method in~\cite{srivastava2015unsupervised} because it does not
  scale to high-resolution videos, but only to small patches.}: a) a baseline
which merely copies the last frame used for conditioning, b) a baseline method
which estimates optical flow~\citep{brox2004high} from two consecutive frames
and extrapolates flow in subsequent frames under the assumption of constant flow
speed, c) an adversarially trained multi-scale CNN~\citep{mathieu2015deep} and
several variants of our proposed approach.

Qualitative comparisons can be seen in the fig.~\ref{prediction_results} and in
the supplementary material\footnote{see:
  \url{https://y0ast.github.io/Transformation-Based-Models-of-Video-Sequences/GenerationBenchmark/}}. The first column on the page
shows the input, the second the ground truth, followed by results from our
model, \cite{mathieu2015deep} and optical flow~\citep{brox2004high}. Note
especially the severe deformations in the last two columns, while our model
keeps the frame recognizable. It produces fairly sharp reconstructions
validating our first hypothesis that errors in the space of transformations
still yield plausible reconstructions (see section~\ref{sec:model}). However it
is also apparent that our approach underestimates movement, which follows
directly from using the MSE criterion. As discussed before, MSE in pixel space
leads to blurry results, however using MSE in transformation space also has some
drawbacks. In practice, the model will predict the average of several likely
transformations, which could lead to an understimation of the true movement.

In order to quantify the generation quality we use the metric described in
section~\ref{sec:test}. We use C3D network \citep{tran2015learning} as the video
action classifier: C3D uses both appearance and temporal information jointly,
and is pre-trained with Sports1M~\citep{karpathy2014large} and fine tuned on UCF
101. Due to the model constraints, we trained only two models, that takes $4$
and $8$ frames as input, respectively.

We evaluate the quality of generation using $4$ (the first four predicted
frames) and the whole set of $8$ predicted frames, for the task of action
classification. At test time, we generate frames from each model under
consideration, and then use them as input to the corresponding C3D network.

Table~\ref{table:ucf101} shows the accuracy of our approach and several
baselines. The best performance is achieved by using ground truth frames, a
result comparable to methods recently appeared in the
literature~\citep{karpathy2014large,tran2015learning}. We see that for
ground truth frames, the number of frames (4 or 8) doesn't make a difference.
There is not much additional temporal or spatial signal provided by having
greater than four frames. Next, we evaluate how much we lose by representing
frames as tiled affine transforms. As the second row shows there is negligible
if any loss of accuracy when using frames reconstructed from the estimated
affine transforms (using the method described in section~\ref{sec:ate}), validating
our assumptions at the beginning of section~\ref{sec:model} on how video sequences
can be represented. The next question is then whether these affine transforms
are predictable at all. The last two rows of Table~\ref{table:ucf101} show that
this is indeed the case, to some extent. The longer the sequence of generated
frames the poorer the performance, since the generation task gets more and more
difficult.

Compared to other methods, our approach performs better than optical flow and
even the more sophisticated multi-scale CNN proposed in~\cite{mathieu2015deep}
while being computationally cheaper. For instance, our method has less than half
a million parameters and requires about $2$G floating point operations to
generate a frame at test time, while the multi-scale CNN
of~\cite{mathieu2015deep} has $25$ times more parameters (not counting the
discriminator used at training time) and it requires more than $100$ times more
floating point operations to generate a single frame.

Finally, we investigate the robustness of the system to its hyper-parameters: a)
choice of patch size, b) number of input frames, and c) number of predicted
frames. The results reported in Table~\ref{table:ucf101_robustness} demonstrate
that the model is overall pretty robust to these choices. Using patch sizes that
are too big makes reconstructions blocky but within each block motion is
coherent. Smaller patch sizes give more flexibility but make the prediction task
harder as well. Mapping into patches of size smaller than $16\times16$ seems a
good choice. Using only $2$ input frames does not seem to provide enough context
to the predictor, but anything above $3$ works equally well. Training for
prediction of the next frame works well, but better results can be achieved by
training to predict several frames in the future, overall when evaluating longer
sequences.

\setlength{\tabcolsep}{4pt}
\begin{table}[bth] \centering
  \caption{Analysis of the robustness to the choice of hyper-parameters, shows
    classification scores compared to reference model. The
    reference model takes $4$ frames as input, predicts one frame, and maps $12
      \times 12$ patches onto $8 \times 8$ patches with stride $4$.}
  \label{table:ucf101_robustness}
  \begin{tabular}{l||c | c | c | c}
    Method                                              & 4 frames & 8 frames \\
    \hline \hline reference                             & 63.57    & 57.32    \\
    \hline Varying patch size                           &          &          \\
    \hspace{.2cm} from $32 \times 32$ to $16 \times 16$ & 61.73    & 53.85    \\
    \hspace{.2cm} from $16 \times 16$ to $8 \times 8$   & 63.75    & 57.18    \\
    \hline Number of input frames                       &          &          \\
    \hspace{.2cm} 2                                     & 63.6     & 57.11    \\
    \hspace{.2cm} 3                                     & 63.8     & 57.4     \\
    \hline Number of predicted frames                   &          &          \\
    \hspace{.2cm} 2                                     & 64.1     & 57.5     \\
    \hspace{.2cm} 4                                     & 64.54    & 57.88
  \end{tabular}
\end{table} \setlength{\tabcolsep}{1.4pt}
\newpage
\section{Conclusions}
\label{sec:conclusions}

In this work, we proposed a new approach to generative modeling of video
sequences. This model does not make any assumption about the spatio-temporal
resolution of video sequences nor about object categories. The key insight of
our approach is to model in the space of transformations as opposed to raw pixel
space. {\em A priori} we lack a good metric to measure how well a frame is
reconstructed under uncertainty due to objects motion in natural scenes.
Uncertainty about object motion and occlusions causes blurry generations when
using MSE in pixel space. Instead, by operating in the space of transformations
we aim at predicting how objects move, and estimation errors only yield a
different, and possibly still plausible, motion. With this motivation we proposed
a simple CNN operating in the space of affine transforms and we showed that it
can generate sensible sequences up to about $4$ frames. This model produces
sequences that are both visually and quantitatively better than previously
proposed approaches.

The second contribution of this work is the metric to compare generative models
of video sequences. A good metric should not penalize a generative model for
producing a sequence which is plausible but different from the ground truth.
With this goal in mind and assuming we have at our disposal labeled sequences,
we can first train a classifier using ground truth sequences. Next, the
classifier is fed with sequences produced by our generative model for evaluation. A
good generative model should produce sequences that still retain discriminative
features. In other words, plausibility of generation is assessed in terms of how
well inherent information is preserved during generation as opposed to
necessarily and merely reproducing the ground truth sequences.

The proposed model is relatively simple; straightforward extensions that could improve
its prediction accuracy are the use of a multi-scale architecture and the
addition of recurrent units. These would enable a better modeling of objects of
different sizes moving at varying speeds and to better capture complex
temporal dynamics (e.g., cyclical movements like walking). A larger extension
would be the addition of an appearance model, which together with our explicit
transformation model could lead to learning better feature representations for
classification.

In our view, the proposed approach should be considered as a stronger baseline
for future research into next frame prediction. Even though our analysis shows
improved performance and better looking generations, there are also obvious
limitations. The first such limitation is the underestimation of transformations
due to usage of the MSE as a criterion. We consider two main avenues worth
pursuing in this space. First, we consider modelling a distribution of
transformations and sampling one from it. The challenge of this approach is to
sample a consistent trajectory. One could model the distribution of an entire
trajectory, but that is a complex optimization problem. A second option is to
use adversarial training to force the model to pick a plausible action. This
option does not guarantee that underestimation of movement will be avoided. This
will depend on the discriminator model accepting this as a plausible option.

Another limitation is that the current model does not factor out the ``what'' from the
``where'', appearance from motion. The representation of two distinct objects
subject to the same motion, as well as the representation of the same object
subject to two different motion patterns are intrinsically different. Instead,
it would be more powerful to learn models that can discover such factorization
and leverage it to produce more efficient and compact representations.

\subsubsection*{Acknowledgments}
Authors thank Camille Couprie and Michael Mathieu for
discussions and helping with evaluation of their models.

\newpage
\bibliography{references}
\bibliographystyle{iclr2017_conference}

\end{document}